\documentclass[10pt,a4paper,conference]{IEEEtran}
\ifCLASSINFOpdf
\else
\fi
\usepackage{amsmath,amsfonts,bm}

\def\eqref#1{equation~\ref{#1}}

\def\1{\bm{1}}

\def\vtheta{{\bm{\theta}}}

\def\vx{{\bm{x}}}
\def\vy{{\bm{y}}}

\DeclareMathAlphabet{\mathsfit}{\encodingdefault}{\sfdefault}{m}{sl}
\SetMathAlphabet{\mathsfit}{bold}{\encodingdefault}{\sfdefault}{bx}{n}

\usepackage{times}

\usepackage{soul}
\usepackage{url}
\usepackage[hidelinks]{hyperref}
\usepackage[utf8]{inputenc}
\usepackage[small]{caption}
\usepackage{graphicx}
\usepackage{amsmath}
\usepackage{booktabs}
\usepackage{algorithm}
\usepackage{algpseudocode}
\usepackage{makecell}

\begin{document}
%

\title{MetaMix: Improved Meta-Learning with Interpolation-based Consistency Regularization}

\makeatletter
\newcommand{\linebreakand}{%
  \end{@IEEEauthorhalign}
  \hfill\mbox{}\par
  \mbox{}\hfill\begin{@IEEEauthorhalign}
}
\makeatother

\author{\IEEEauthorblockN{Yangbin Chen}
\IEEEauthorblockA{Department of Computer Science\\
City University of Hong Kong\\
Email: robinchen2-c@my.cityu.edu.hk}
\and
\IEEEauthorblockN{Yun Ma}
\IEEEauthorblockA{Department of Computing\\
The Hong Kong Polytechnic University\\
Email: mayun371@gmail.com}
\and
\IEEEauthorblockN{Tom Ko}
\IEEEauthorblockA{Department of Computer Science\\ and Engineering\\
Southern University of Science and Technology\\
Email: tomkocse@gmail.com}
\linebreakand
\IEEEauthorblockN{Jianping Wang}
\IEEEauthorblockA{Department of Computer Science\\
City University of Hong Kong\\
Email: jianwang@cityu.edu.hk}
\and
\IEEEauthorblockN{Qing Li}
\IEEEauthorblockA{Department of Computing\\
The Hong Kong Polytechnic University\\
Email: qing-prof.li@polyu.edu.hk}
}


%


\maketitle

\begin{abstract}
Model-Agnostic Meta-Learning (MAML) and its variants are popular few-shot classification methods.
They train an initializer across a variety of sampled learning tasks (also known as episodes) such that the initialized model can adapt quickly to new ones.
However, current MAML-based algorithms have limitations in forming generalizable decision boundaries.
In this paper, we propose an approach called MetaMix.
It generates virtual feature-target pairs within each episode to regularize the backbone models.
MetaMix can be integrated with any of the MAML-based algorithms and learn the decision boundaries generalizing better to new tasks.
Experiments on the \textit{mini}-ImageNet, CUB, and FC100 datasets show that MetaMix improves the performance of MAML-based algorithms and achieves state-of-the-art result when integrated with Meta-Transfer Learning.
\end{abstract}


%
\IEEEpeerreviewmaketitle

\section{Introduction}
\label{sec1:intro}

Deep learning methods have achieved remarkable success in the field of computer vision, speech recognition, and natural language understanding.
However, the impressive performance relies much on training a deep neural network with large-scale human-annotated data.
It often struggles to generalize to new tasks with limited annotated data, while human beings are capable of learning new tasks rapidly by utilizing what they learned in the past \cite{wang2020generalizing}.

Few-shot classification methods aim to train a classifier with limited training examples for the novel classes, using knowledge learned from previous classes \cite{fei2006one,lake2015human}.
Recently, meta-learning (also called `learning to learn'), which intends to make a rapid adaptation to new environments with a few examples, has been applied to tackle the few-shot classification problems \cite{chen2018investigation,ko2020prototypical,Liu_2020_CVPR}.

Model-Agnostic Meta-Learning (MAML) \cite{finn2017model} has been one of the most successful meta-learning algorithms.
The MAML algorithm trains an initializer across various sampled learning tasks (episodes) so that the initialized model can adapt quickly to new tasks with only a few labeled examples.
But the learned model is prone to overfitting by memorizing the training data, and hence difficult to generalize to unseen data.

To improve the model generalization ability with a few examples, previous meta-learning work like \cite{zhang2018metagan} aims to learn a sharper decision boundary by adversarial training within each episode.
\cite{rusu2018meta} learned a data-dependent latent generative representation of model parameters and performed gradient-based meta-learning in the low-dimensional latent space.


In this paper, we propose a framework, MetaMix, which aims to regularize the training process of the backbone models.
It has a low computational cost and is stable to train the backbone models.
Specifically, MetaMix produces virtual examples within each episode using an interpolation-based method called \textit{mixup} \cite{zhang2018mixup}.
It is a simple but effective method to generate examples by linear interpolations of the training feature vectors and their corresponding labels.
Utilizing such virtual feature-target vectors encourages a smoother model behavior along the data space, avoiding sudden oscillations around the training examples.

We integrate MetaMix with four representative MAML-based algorithms: the original MAML, First-Order MAML (FOMAML) \cite{nichol2018first}, Meta-SGD \cite{li2017meta}, and Meta-Transfer Learning (MTL) \cite{sun2019meta}.
Experiments show that MetaMix improves the performance of all four MAML-based algorithms on three popular few-shot image classification datasets.
Furthermore, it achieves state-of-the-art results when integrated with MTL.

Our contributions in this paper are summarized as follows:
\begin{itemize}
    \item We propose MetaMix as an approach, which can be integrated with many meta-learning algorithms, including MAML and its variants, and improve their performance.
    \item We find that MAML-based algorithms with MetaMix achieve comparable performance even if using only 50\% of the training data.
    \item MetaMix with MTL achieves state-of-the-art results.
\end{itemize}

In the coming sections, we review works related to Few-Shot Learning (FSL), meta-learning, and \textit{mixup} training in Section \ref{sec2:preliminaries}.
In Section \ref{sec3:method}, we introduce the details of our MetaMix approach.
In Section \ref{sec4:exp}, we present the experiments and analyze the results.
Conclusions are drawn in Section \ref{sec5:conc}.

\section{Preliminaries}
\label{sec2:preliminaries}

\subsection{Few-shot learning}
\label{subsec2.1:fsl}

Few-shot learning is a machine learning problem in which only a few examples with supervised information can be used to train a model.
\cite{wang2020generalizing} provides a reasonable interpretation to tackle few-shot learning problems, inspired by error decomposition in supervised learning.
According to their description, finding an optimal hypothesis function $\hat{h}=argmin_hR(h)$ of a learning problem, where $R$ is the expected risk, can be decomposed into two works.
One is to find a hypothesis space $\mathcal{H}$ which contains the best approximation function $h^*=argmin_{h \in \mathcal{H}}R(h)$ for $\hat{h}$.
The other is to find the best hypothesis function $h_I = argmin_{h \in \mathcal{H}}R_I(h)$ in $\mathcal{H}$ by empirical risk minimization, where $R_I$ is the empirical risk.
In this way, the error can be formulated as:
\begin{equation}
    \mathbb{E}[R(h_I)-R(\hat{h})]=\mathbb{E}[R(h^*)-R(\hat{h})] + \mathbb{E}[R(h_I)-R(h^*)]
\end{equation}
On the right side of the equation, the first part is the {\it approximate error} measuring how close the functions in $\mathcal{H}$ can approximate $\hat{h}$ and the second part is the {\it estimate error} measuring the effect of empirical risk minimization.
When labeled examples in a task are limited, purely using the empirical risk minimization on the examples can be unreliable.
Consequently, the keys to solve few-shot learning problems by incorporating prior knowledge include: (1) augmenting data to reduce the estimate error, (2) regularizing the model to constrain the complexity of $\mathcal{H}$ and reduce the approximate error, (3) making a good search for $\vtheta$ parameterizing the best hypothesis $h^*$ to reduce the estimate error.

\subsection{Meta-learning for few-shot classification}
\label{subsec2.2:metalearning}

Few-shot classification is a few-shot learning problem in classification tasks.
Meta-learning has attracted many researchers' attention to solve few-shot classification problems.
A conventional way to train a classification model usually includes several steps: designing a model architecture, choosing an initializer, feeding training examples in terms of mini-batches, computing the loss, and updating the parameters with a proper optimizer via backpropagation.
It treats the training examples as basic units.
Meta-learning, distinctively, considers the entire classification task itself as a basic unit.
It trains the model across different tasks so that the steps mentioned above are possible to be learned, for example, learning to initialize the parameters \cite{finn2017model} and learning to design an optimization algorithm \cite{andrychowicz2016learning}.
That is why meta-learning is also known as `learning to learn'.

There are three types of common meta-learning algorithms which can be explained by the keys mentioned in section \ref{subsec2.1:fsl}.
The metric-based algorithms learn a shared metric model between examples from the query set and the support set \cite{snell2017prototypical,sung2018learning}.
It is a kind of task-invariant embedding learning strategy, which constrains a smaller $\mathcal{H}$.
The model-based algorithms incorporate external memory to the model architectures or design the training process for rapid generalization \cite{santoro2016meta,munkhdalai2017meta}, which reduces the size of $\mathcal{H}$.
The optimization-based algorithms focus on improving the gradient-based optimization algorithms for learning better with a few examples \cite{finn2017model,Ravi2017OptimizationAA}, which is to search good $\vtheta$ to parametrize $h^*$.

MAML is one of the most popular optimization-based meta-learning algorithms.
It trains an initializer across various sampled episodes so that the initialized model can adapt quickly to new tasks.
Specifically, a mini-batch of episodes starts with the same initializer.
Two optimization loops are playing different roles in a mini-batch.
Within each episode, the inner loop updates the backbone model using the support set, and the outer loop computes the loss on the query set using the updated backbone model.
After that, a meta-initializer is learned by a combination of the losses of the outer loop, across the mini-batch of episodes (with second-order gradient updates).
In this way, the inner loop optimization is guided by the outer loop meta-objective.
MAML is followed by a number of research works \cite{nichol2018first,li2017meta,sun2019meta}.
In this paper, we choose to improve the performance of MAML and its three variants by introducing our MetaMix approach (see Figure
\ref{fig1:framework}).

\subsection{\textit{mixup} Training}
\label{subsec2.3:mixup}

The \textit{mixup} method trains a neural network on convex combinations of pairs of examples and their labels \cite{zhang2018mixup}.
It acts as a regularization technique to improve the generalization of the neural architectures by producing virtual feature-target vectors from the vicinity distribution of the training examples.
Neural networks trained with \textit{mixup}, which corresponds to the Vicinal Risk Minimization (VRM), provides a smoother estimate of uncertainty, compared with those with Empirical Risk Minimization (ERM), which allows the networks to memorize the training data.

The \textit{mixup} training method has many following works.
Some works extend the method itself.
\cite{verma2018manifold} mixed on random hidden layer representations.
\cite{guo2019mixup} proposed the adaptive generation of the mixing ratio for a specific pair to avoid overlapping between the mixed examples and the real ones.
Some works integrate the method with other algorithms.
In semi-supervised learning, \cite{verma2019interpolation} substituted the labels by soft labels from a teacher model.
\cite{ma2019mixing} proposed a generalized framework, which did {\it mixup} between the training examples and adversarial examples.
Some holistic methods in semi-supervised learning combine {\it mixup} with other techniques such as multiple data augmentation and label sharpening to obtain strong empirical results \cite{berthelot2019mixmatch,berthelot2019remixmatch,sohn2020fixmatch}.
Furthermore, \cite{mao2019virtual} applied {\it mixup} to unsupervised domain adaptation by mixing on the logits of the unlabeled input data.

Our work incorporates \textit{mixup} to meta-learning algorithms, which can also be explained by the keys in section \ref{subsec2.1:fsl}.
Traditional meta-learning algorithms may face meta-overfitting problems, which learn knowledge resulting in a hypothesis space too tightly around solutions to the source tasks \cite{hospedales2020meta}.
{\it mixup} performs as a regularizer, which prevents meta-overfitting and constrains the hypothesis space generalizing more to the target task (key (2)).
It is orthogonal to MAML-based algorithms (key (3)).

\begin{figure*}[h]
\centering
\includegraphics[width=16cm]{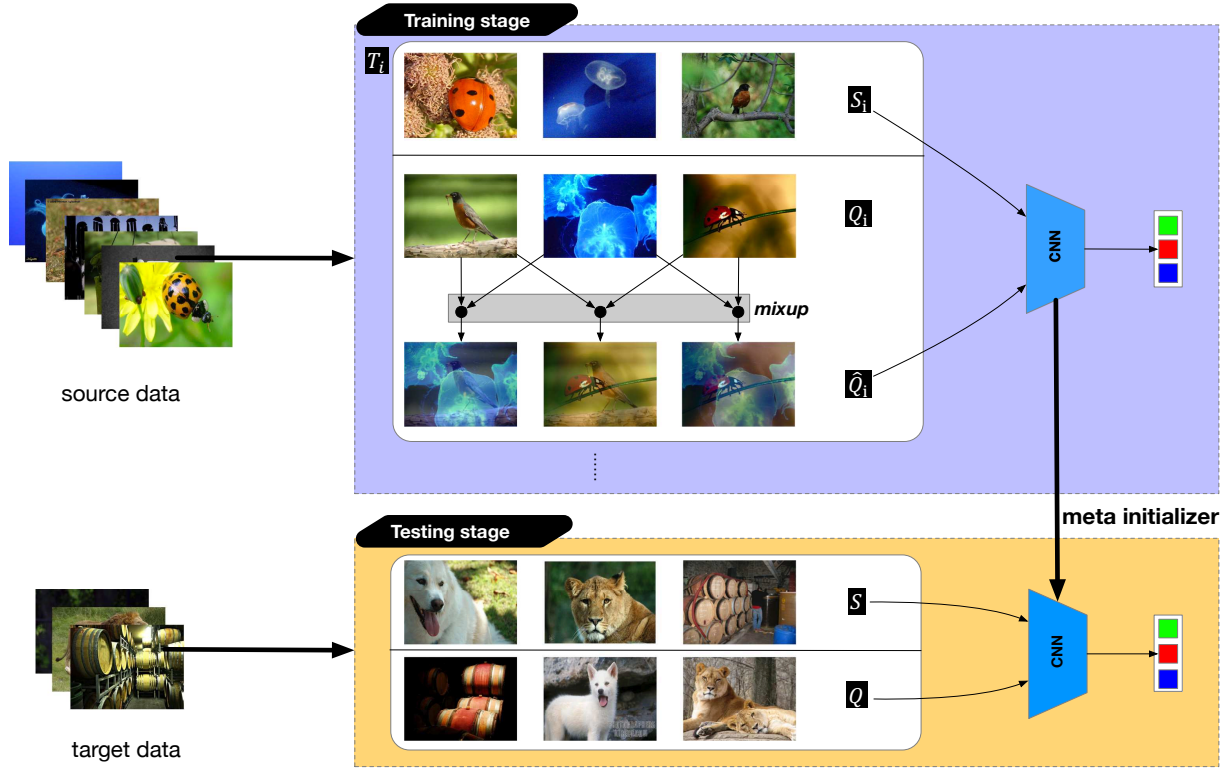}
\caption{{\bf Framework of MetaMix with MAML and its variants.} We take a 3-way, 1-shot task as an example. In the training stage, a set of episodes are sampled from the training set. Within each episode $\mathcal{T}_i$, we build a support set $\mathcal{S}_i$ and a query set $\mathcal{Q}_i$. MetaMix generates virtual examples using \textit{mixup} and gets a new query set $\hat{\mathcal{Q}}_i$. $\mathcal{S}_i$ and $\hat{\mathcal{Q}}_i$ are used to train an initializer. In the testing stage, we sample episodes from novel classes. We fine-tune the initialized model through the support set $\mathcal{S}$ and evaluate the updated model by the query set $Q$.}
\label{fig1:framework}
\end{figure*}

\section{Methodology}
\label{sec3:method}

\subsection{Problem definition}
\label{subsec3.1:define}

A few-shot classification task, also named as $N$-way, $K$-shot, is defined as follows.
Given a training set $\mathcal{X}$ containing many labeled examples per class from $M$ classes, and a test set $\mathcal{V}$ containing $K$ labeled examples per class from $N$ novel classes, while $K$ is small, a few-shot classifier is learned to recognize the $N$ novel classes.
Many meta-learning algorithms train the classification model on a set of episodes $\{\mathcal{T}_i\}_{i=1}^B$, which are sampled from $\mathcal{X}$, according to a distribution $p(\mathcal{T})$.
Similar to the target task, the episode $\mathcal{T}_i$ is an $N$-way classification task consisting of a support set $\mathcal{S}_i$ and a query set $\mathcal{Q}_i$.
$\{\mathcal{S}_i, \mathcal{Q}_i\}_{i=1}^B$ is used to train the model in a supervised way.

\subsection{Motivation}
\label{subsec:motivation}

MAML and its variants contain two levels of optimization, in which the outer loop optimization with a meta-objective guides the inner loop optimization.
In this way, they can produce model initializers generalizing well to any new task.
Our proposed method aims to regularize the training process of the outer loop to improve the generalization of the model initializers learned by MAML and its variants.

\subsection{MetaMix with MAML}
\label{subsec3.2:algorithm}

The detailed algorithm of MetaMix with MAML is described in Algorithm \ref{alg:metamix}.
In each training iteration, the model is trained on a set of episodes.
When the target task is $N$-way, $K$-shot, the sampled episodes are usually $N$-way tasks, with randomly selected $N$ classes for each episode.
An episode $\mathcal{T}_i$ consists of a support set $\mathcal{S}_i=\{(\vx_j,\vy_j);j \in (1,2,...,J)\}$ and a query set $\mathcal{Q}_i=\{(\vx_z,\vy_z);z \in (1,2,...,Z)\}$.
We usually set $J$ to $N \times K$ and $Z$ to $N \times H$.

Considering a backbone model represented by a function $f(\vtheta)$ with parameters $\vtheta$, firstly we use the support set to update the parameters for an efficient adaptation to $\mathcal{T}_i$.
A cross-entropy loss function $\mathcal{L}_{\mathcal{S}_i}$ is defined to update the parameters through one or more gradient descent steps:
\begin{equation}
\label{eq:1}
 \mathcal{L}_{\mathcal{S}_i}(f_\vtheta)  = -\sum_{(\vx_j,\vy_j) \in \mathcal{S}_i} \vy_j log f_{\vtheta}(\vx_j)
\end{equation}
A one-step gradient descent is as follows:
\begin{equation}
\label{eq:2}
 \vtheta'_i = \vtheta - \alpha \cdot \nabla_{\vtheta}\mathcal{L}_{\mathcal{S}_i}(f_\vtheta)
\end{equation}
where $\alpha$ is the learning rate that can be fixed like in MAML or be learned like in MetaSGD.
It is also called the inner loop, which adapts to each separate task through a few gradient updates.

Then the parameters are optimized on the performance of $f(\vtheta'_i)$ evaluated by the query set.
In our work, we sample any two examples $(\vx_m,\vy_m)$ and $(\vx_n,\vy_n)$ from the query set, and compute a linear interpolation of both their features and labels:
\begin{equation}
\label{eq:3}
    \hat{\vx}_z = \lambda \vx_m + (1-\lambda) \vx_n
\end{equation}
\begin{equation}
\label{eq:4}
    \hat{\vy}_z = \lambda \vy_m + (1-\lambda) \vy_n
\end{equation}
where $\lambda$ is an interpolation coefficient generated from a Beta distribution $\lambda \sim \bf{B}(\check{\alpha},\check{\alpha})$.
We replace the original examples with the generated virtual examples, and get a new query set $\hat{\mathcal{Q}}_i=\{(\hat{\vx}_z,\hat{\vy}_z);z \in (1,2,...,Z)\}$.
$\mathcal{L}_{\mathcal{Q}_i}(f_{\vtheta'_i})$ is the cross-entropy loss over the new query set examples:
\begin{equation}
\label{eq:5}
    \mathcal{L}_{\hat{\mathcal{Q}}_i}(f_{\vtheta'_i}) = - \sum_{(\hat{\vx}_z,\hat{\vy}_z) \in \hat{\mathcal{Q}}_i} \hat{\vy}_z log f_{\vtheta'_i}(\hat{\vx}_z)
\end{equation}
The parameters of the backbone model are trained by minimizing the total loss across the episodes from the batch:
\begin{equation}
\label{eq:6}
    \vtheta \leftarrow \vtheta - \beta \cdot \nabla_{\vtheta} \sum_{i}\mathcal{L}_{\hat{\mathcal{Q}}_i}(f_{\vtheta'_i})
\end{equation}
where $\beta$ is the learning rate.
It is also called the outer loop, which initializes the parameters using one or more episodes in every training iteration, enabling fast adaptations to new tasks.

In our work, we generate virtual examples from the query set for two reasons.
One reason is that the query set is responsible for optimizing the meta-objective across different episodes, which is significant to the generalization of the learned initializer.
The second reason is that within an episode, the query set usually contains more examples than the support set.
Virtual examples generated by interpolating examples from the query set get more similar to real data distribution.
We do an ablation study to compare the effectiveness of mixing on different sets in section \ref{subsec4.3:results}.

There are works augmenting
the training set within each episode like \cite{chen2019image}.
It learns to combine training samples.
But they generate deformed images by interpolating an image from the support set and an image randomly sampled from the entire source data.
Moreover, they assign the class label of the support set image to the synthesized deformed image, which is purely a data augmentation method rather than a regularization technique.

\begin{algorithm}[h]
\caption{MetaMix with MAML}
\label{alg:metamix}
\begin{algorithmic}[1]
    \Require
        $p(\mathcal{T}):$ distribution over tasks
    \Require
        $\mathcal{S}_i:$ support set; $\mathcal{Q}_i:$ query set
    \Require
        $\alpha, \beta:$ learning rate
    \Require
        $\check{\alpha}:$ Beta distribution parameter
    \Require
        $mix_{\lambda}(a,b) = \lambda a + (1-\lambda) b, \lambda \sim {\bf B}(\check \alpha, \check \alpha)$
        
\State Randomly initialize model parameters $\vtheta$
\While{not done}
    \State Sample a batch of episodes $\mathcal{T}_i \sim p(\mathcal{T})$
    \ForAll{$\mathcal{T}_i$}
        \State Sample a support set $\mathcal{S}_i=\{(\vx_j,\vy_j)\}_{j=1}^J$
        \State Evaluate $\nabla_{\vtheta}\mathcal{L}_{\mathcal{S}_i}(f_\vtheta)$ using $\mathcal{S}_i$ and $\mathcal{L}_{\mathcal{S}_i}(f_\vtheta)$
        \State Compute adapted parameters with gradient descent: $\vtheta'_i = \vtheta - \alpha \cdot \nabla_{\vtheta}\mathcal{L}_{\mathcal{S}_i}(f_\vtheta)$
        \State Sample a query set $\mathcal{Q}_i=\{(\vx_z,\vy_z)\}_{z=1}^Z$
        \State Randomly select pairs of examples $\{(\vx_m,\vy_m)\}_{m=1}^Z,\{(\vx_n,\vy_n)\}_{n=1}^Z$ from $\mathcal{Q}_i$
        \State $\hat{\vx}_z=mix_{\lambda}(\vx_m,\vx_n), \hat{\vy}_z=mix_{\lambda}(\vy_m,\vy_n)$
        \State Get new query set $\hat{\mathcal{Q}}_i=\{(\hat{\vx}_z,\hat{\vy}_z)\}_{z=1}^Z$
    \EndFor
    \State Update $\vtheta \leftarrow \vtheta - \beta \cdot \nabla_{\theta} \sum_{i}\mathcal{L}_{\hat{\mathcal{Q}}_i}(f_{\vtheta'_i})$
\EndWhile
\end{algorithmic}
\end{algorithm}

\subsection{MetaMix with MAML variants}
\label{subsec3.3:algorithm1}

We also integrate MetaMix with three MAML variants.
The FOMAML algorithm omits the second derivatives of MAML, which are calculated in the outer loop.
It reduces the computational cost while the performance does not drop too much.
The Meta-SGD algorithm not only learns to learn the learner's initialization but also the learner's update direction and the learning rate, which has a much higher capacity compared to MAML.

Both FOMAML and Meta-SGD use the same backbone model as the original MAML does, which is a shallow CNN with 4 CONV layers.
However, the MTL incorporates ResNet-12 \cite{he2016deep}, a much deeper neural network, as the backbone model.
There are two stages to train the model.
Firstly, the ResNet-12 model is pre-trained over the entire training set to learn a good representation.
After that, the output layer is removed, and the rest of the parameters are frozen.
Secondly, the pre-trained model is transferred to the meta-learning stage, added new output layers.
The new backbone model is trained by MAML, in which only parameters of the new output layers will be updated.
MTL achieves much better performance than the other algorithms do.

In our work, we integrate MetaMix with FOMAMAL and Meta-SGD in the same way as with MAML.
As for the MTL algorithm, first of all, we generate virtual examples from the entire training set and use the virtual examples to pre-train the model.
Then we do the same training as the other algorithms during the meta-learning stage.

\section{Experiments}
\label{sec4:exp}

\subsection{Datasets}
\label{subsec4.1:dataset}

\textbf{\textit{mini}-ImageNet} is a dataset from the ILSVRC-2015 \cite{russakovsky2015imagenet}.
It contains 100 classes of color images.
Each class has 600 images, and the images are resized to 84 $\times$ 84.
Following \cite{Ravi2017OptimizationAA}, we split the 100 classes into three parts: 64 for training, 16 for validation, and 20 for testing.

\textbf{Caltech-UCSD Birds-200-2011} (referred to CUB hereafter) is a dataset for bird species classification \cite{wah2011caltech}.
It has 200 classes of total 11,788 images, which are resized to 84 $\times$ 84.
We randomly choose 100 for training, 50 for validation, and 50 for testing classes, following the work of \cite{hilliard2018few}. 

\textbf{Fewshot-CIFAR100} (referred to FC100 hereafter) is a dataset from the object classification dataset CIFAR100 \cite{krizhevsky2009learning}.
It contains 100 classes of color images.
Each class has 600 images of 32 $\times$ 32.
The tasks on FC100 is more difficult since the image resolution is lower.
To split the dataset, we follow the work of \cite{oreshkin2018tadam}, which separates the classes according to the object super-classes.
The training set is from 60 classes belonging to 12 super-classes.
The validation and testing sets are from 20 classes belonging to 4 super-classes, respectively.

\subsection{Implementation details}
\label{subsec4.2:implementation}

\noindent
\textbf{Sampled episodes} \space For an $N$-way, $K$-shot classification task, a set of episodes are sampled from the training, validation, and testing sets.
Within each episode, we randomly select $N$ classes to build a support set and a query set.
The support set consists of $K$ examples per class, and the query set consists of $H$ examples per class.
In our experiment, $N$ is 5, $K$ is 1 or 5, and $H$ is 16.
In the training stage, algorithms except MTL and MetaMix with MTL train 60,000 episodes for the 5-way, 1-shot task and 40,000 episodes for the 5-way, 5-shot task.
The MTL and MetaMix with MTL algorithms train only 8,000 episodes because they have pre-trained a ResNet-12 model over the entire training set for 10,000 iterations.
In the validation and testing stages, we fine-tune the learned initializer using examples from the support set and evaluate it with examples from the query set.
All the algorithms are evaluated by the averaged result of 600 episodes, which are randomly sampled from the validation and test sets, respectively.
The result of the validation episodes is used to select the training iteration with the best accuracy, and the result of the testing episodes is used to evaluate the meta-learning algorithms.

\noindent
\textbf{Backbone models} \space Algorithms except for MTL and MetaMix with MTL use a shallow CNN architecture, with four-layer convolutional blocks as the backbone model.
Each block has 32 3 $\times$ 3 filters, followed by ReLU, batch normalization, and a 2 $\times$ 2 max-pooling layer.
MTL and MTL with MetaMix freeze the parameters of the pre-trained ResNet-12 model and add new fully-connected layers for the few-shot tasks so that only parameters of these layers will be updated.

\noindent
\textbf{Important hyperparameters} \space We set the batch size to 4 for the 5-way, 1-shot and 5-way, 5-shot classification tasks.
The learning rate $\alpha$ of the inner loop is set to 0.01 and the outer loop uses Adam as the optimizer.
The mixup hyperparameter $\check{\alpha}$ for Beta distribution is set to 1.0.

\noindent
\textbf{Implementation codes} \space We follow the implementations of \cite{chen2018a} in both the experimental settings and the codes \footnote{https://github.com/wyharveychen/CloserLookFewShot} for all algorithms except Meta-SGD, MetaMix with Meta-SGD, MTL, and MetaMix with MTL.
Among the four exceptions, for the first two algorithms, we refer to an open-source implementation \footnote{https://github.com/jik0730/Meta-SGD-pytorch}, and for the last two algorithms, we refer to an officially released implementation \footnote{https://github.com/yaoyao-liu/meta-transfer-learning}.
All the experiments are based on PyTorch\cite{paszke2017automatic}.

\begin{table*}[h]
\begin{center}
\begin{tabular}{lllllll}
\Xhline{2\arrayrulewidth}
& \multicolumn{2}{c}{\textit{mini}-ImageNet} & \multicolumn{2}{c}{CUB} & \multicolumn{2}{c}{FC100} \\
\multicolumn{1}{c}{\bf Models} &\multicolumn{1}{c}{1-shot} &\multicolumn{1}{c}{5-shot}  &\multicolumn{1}{c}{1-shot} &\multicolumn{1}{c}{5-shot}  &\multicolumn{1}{c}{1-shot} &\multicolumn{1}{c}{5-shot} \\
\hline \hline
\multicolumn{1}{l}{Matching Network} & \multicolumn{1}{c}{ 50.47 $\pm$ 0.80 } & 
\multicolumn{1}{c}{ 64.83 $\pm$ 0.67 } & \multicolumn{1}{c}{ 57.70 $\pm$ 0.87 } & 
\multicolumn{1}{c}{ 71.42 $\pm$ 0.71 } & \multicolumn{1}{c}{ 36.97 $\pm$ 0.67 } & 
\multicolumn{1}{c}{ 49.44 $\pm$ 0.71 }\\
\multicolumn{1}{l}{Prototypical Network} & \multicolumn{1}{c}{ 49.33 $\pm$ 0.82 } & \multicolumn{1}{c}{ 65.71 $\pm$ 0.67 } & \multicolumn{1}{c}{ 51.34 $\pm$ 0.86 } &
\multicolumn{1}{c}{ 67.56 $\pm$ 0.76 } & \multicolumn{1}{c}{ 36.83 $\pm$ 0.69 } &
\multicolumn{1}{c}{ 51.21 $\pm$ 0.74 }\\
\multicolumn{1}{l}{Relation Network} & \multicolumn{1}{c}{ 50.48 $\pm$ 0.80 } &
\multicolumn{1}{c}{ 65.39 $\pm$ 0.72 } & \multicolumn{1}{c}{ 59.47 $\pm$ 0.96 } &
\multicolumn{1}{c}{ 73.88 $\pm$ 0.74 } & \multicolumn{1}{c}{ 36.40 $\pm$ 0.69 } &
\multicolumn{1}{c}{ 51.35 $\pm$ 0.69 } \\
\hline
\multicolumn{1}{l}{MAML} & \multicolumn{1}{c}{48.18 $\pm$ 0.78} & \multicolumn{1}{c}{ 63.05 $\pm$ 0.71} & \multicolumn{1}{c}{ 54.32 $\pm$ 0.91} & \multicolumn{1}{c}{ 71.37 $\pm$ 0.76} & \multicolumn{1}{c}{ 35.96 $\pm$ 0.71} & \multicolumn{1}{c}{ 48.06 $\pm$ 0.73}\\
\multicolumn{1}{l}{MetaMix+MAML}  & \multicolumn{1}{c}{\bf 50.51 $\pm$ 0.86} & \multicolumn{1}{c}{\bf 65.73 $\pm$ 0.72} & \multicolumn{1}{c}{\bf 57.70 $\pm$ 0.92} & \multicolumn{1}{c}{\bf 73.66 $\pm$ 0.74} & \multicolumn{1}{c}{\bf 37.09 $\pm$ 0.74} & \multicolumn{1}{c}{\bf 49.31 $\pm$ 0.72}\\
\hline
\multicolumn{1}{l}{FOMAML} & \multicolumn{1}{c}{45.22 $\pm$ 0.77} & \multicolumn{1}{c}{60.97 $\pm$ 0.70} & \multicolumn{1}{c}{53.12 $\pm$ 0.93} & \multicolumn{1}{c}{70.90 $\pm$ 0.75} & \multicolumn{1}{c}{34.97 $\pm$ 0.70} & \multicolumn{1}{c}{47.41 $\pm$ 0.73} \\
\multicolumn{1}{l}{MetaMix+FOMAML} & \multicolumn{1}{c}{\bf 47.78 $\pm$ 0.77} & \multicolumn{1}{c}{\bf 63.55 $\pm$ 0.70} & \multicolumn{1}{c}{\bf 54.81 $\pm$ 0.97} & \multicolumn{1}{c}{\bf 72.90 $\pm$ 0.74} & \multicolumn{1}{c}{\bf 36.48 $\pm$ 0.67} & \multicolumn{1}{c}{\bf 49.48 $\pm$ 0.71}\\
\hline
\multicolumn{1}{l}{MetaSGD} & \multicolumn{1}{c}{ 49.93 $\pm$ 1.73 } &
\multicolumn{1}{c}{ 64.01 $\pm$ 0.90 } & \multicolumn{1}{c}{ 56.19 $\pm$ 0.92 } &
\multicolumn{1}{c}{ 69.14 $\pm$ 0.75 } & \multicolumn{1}{c}{ 36.36 $\pm$ 0.66 } &
\multicolumn{1}{c}{ 49.96 $\pm$ 0.72 }\\
\multicolumn{1}{l}{MetaMix+MetaSGD} & \multicolumn{1}{c}{\bf 50.60 $\pm$ 1.80 } & \multicolumn{1}{c}{\bf 64.47 $\pm$ 0.88 } & \multicolumn{1}{c}{\bf 57.64 $\pm$ 0.88 } & \multicolumn{1}{c}{\bf 70.50 $\pm$ 0.70 } & \multicolumn{1}{c}{\bf 37.44 $\pm$ 0.71 } & \multicolumn{1}{c}{\bf 51.41 $\pm$ 0.69 } \\
\hline
\multicolumn{1}{l}{MTL} & \multicolumn{1}{c}{61.37 $\pm$ 0.82} & 
\multicolumn{1}{c}{ 78.37 $\pm$ 0.60} & \multicolumn{1}{c}{71.90 $\pm$ 0.86} & \multicolumn{1}{c}{ 84.68 $\pm$ 0.53} & \multicolumn{1}{c}{42.17 $\pm$ 0.79} & \multicolumn{1}{c}{ 56.84 $\pm$ 0.75}\\
\multicolumn{1}{l}{MetaMix+MTL} & \multicolumn{1}{c}{\bf 62.74 $\pm$ 0.82} & \multicolumn{1}{c}{\bf 79.11 $\pm$ 0.58} & \multicolumn{1}{c}{\bf 73.04 $\pm$ 0.86} & \multicolumn{1}{c}{\bf 86.10 $\pm$ 0.50} & \multicolumn{1}{c}{\bf 43.58 $\pm$ 0.73} & \multicolumn{1}{c}{\bf 58.27 $\pm$ 0.73}\\
\Xhline{2\arrayrulewidth}
\end{tabular}
\end{center}
\caption{ Accuracy with 95\% confidence intervals of \textbf{5-way, $K$-shot ($K$=1, 5)} classification tasks on \textbf{\textit{mini}-ImageNet}, \textbf{CUB}, and \textbf{FC100} datasets.  }
\label{tab1:supervised}
\end{table*}

\begin{table*}[h]
\begin{center}
\begin{tabular}{lllllll}
\Xhline{2\arrayrulewidth}
& \multicolumn{2}{c}{\textit{mini}-ImageNet} & \multicolumn{2}{c}{CUB} & \multicolumn{2}{c}{FC100} \\
\multicolumn{1}{c}{\bf Models} &\multicolumn{1}{c}{1-shot} &\multicolumn{1}{c}{5-shot}  &\multicolumn{1}{c}{1-shot} &\multicolumn{1}{c}{5-shot}  &\multicolumn{1}{c}{1-shot} &\multicolumn{1}{c}{5-shot} \\
\hline \hline

\multicolumn{1}{l}{MAML(100\%)} & \multicolumn{1}{c}{48.18 $\pm$ 0.78} & \multicolumn{1}{c}{ 63.05 $\pm$ 0.71} & \multicolumn{1}{c}{ 54.32 $\pm$ 0.91} & \multicolumn{1}{c}{ 71.37 $\pm$ 0.76} & \multicolumn{1}{c}{ 35.96 $\pm$ 0.71} & \multicolumn{1}{c}{ 48.06 $\pm$ 0.73}\\
\multicolumn{1}{l}{MetaMix+MAML(100\%)}  & \multicolumn{1}{c}{\bf 50.51 $\pm$ 0.86} & \multicolumn{1}{c}{\bf 65.73 $\pm$ 0.72} & \multicolumn{1}{c}{\bf 57.70 $\pm$ 0.92} & \multicolumn{1}{c}{\bf 73.66 $\pm$ 0.74} & \multicolumn{1}{c}{\bf 37.09 $\pm$ 0.74} & \multicolumn{1}{c}{\bf 49.31 $\pm$ 0.72}\\
\hline
\multicolumn{1}{l}{MAML(50\%)} & \multicolumn{1}{c}{ 46.34 $\pm$ 0.82} & \multicolumn{1}{c}{ 60.47 $\pm$ 0.73} & \multicolumn{1}{c}{ 50.78 $\pm$ 0.86} & \multicolumn{1}{c}{ 65.60 $\pm$ 0.81} & \multicolumn{1}{c}{ 35.38 $\pm$ 0.71} & \multicolumn{1}{c}{ 47.93 $\pm$ 0.78 }\\
\multicolumn{1}{l}{MetaMix+MAML(50\%)} & \multicolumn{1}{c}{\bf 48.04 $\pm$ 0.79} & \multicolumn{1}{c}{\bf 63.52 $\pm$ 0.67} & \multicolumn{1}{c}{\bf 53.22 $\pm$ 0.91} & \multicolumn{1}{c}{\bf 70.13 $\pm$ 0.70} & \multicolumn{1}{c}{\bf 36.35  $\pm$ 0.74 } & \multicolumn{1}{c}{\bf 48.11 $\pm$ 0.69 } \\
\Xhline{2\arrayrulewidth}
\end{tabular}
\end{center}
\caption{ A comparison between using 100\% and 50\% training data; accuracy with 95\% confidence intervals of \textbf{5-way, $K$-shot ($K$=1, 5)} classification tasks on \textbf{\textit{mini}-ImageNet}, \textbf{CUB}, and \textbf{FC100} datasets.}
\label{tab2:labelrate}
\end{table*}

\subsection{Results and analysis}
\label{subsec4.3:results}
\medskip
\noindent
\textbf{Evaluation of MetaMix}
\smallskip

\noindent
In addition to four MAML-based algorithms, we take Matching Network \cite{vinyals2016matching}, Prototypical Network \cite{snell2017prototypical}, and Relation Network \cite{sung2018learning} as baseline algorithms.
Results are given in Table \ref{tab1:supervised}.

We can see that MetaMix improves the performance of all the MAML-based algorithms over three datasets in the 5-way, 1-shot and 5-way, 5-shot classification tasks; meanwhile, MetaMix with MTL achieves state-of-the-art performance.

Specifically, MAML and FOMAML have the most improvement when integrated with MetaMix, in both 1-shot and 5-shot tasks, especially on the \textit{mini}-ImageNet and CUB datasets.
To compare MAML and FOMAML, we can see that although MAML performs better than FOMAML in all tasks, MetaMix with FOMAML achieves comparable performance to MetaMix with MAML in the 5-shot task on CUB and FC100 datasets.
Meta-SGD has the least improvement on \textit{mini}-ImageNet while gets 1.5\% improvement on the other datasets. 
The absolute performance of MetaMix with MTL is much better than the other algorithms because of the deeper network it uses.
As the ResNet model has been a good representation learner, which generalizes well to different kinds of tasks, the improvement of MetaMix is not very obvious, which is around 1.5\%.

Moreover, comparing the results among the three datasets, we find that the FC100 dataset is the most challenging one to do few-shot classification tasks.
The reason is that the images it provides are with low resolution, and the training and testing data are separated according to the object super-classes.
For example, we train the initializer over the images of flowers, but we fine-tune and evaluate the model over the images of animals, which is much more difficult.

\medskip
\noindent
\textbf{Effect of the hyperparameter of Beta distribution}
\smallskip

\noindent
In our experiment, we set the hyperparameter $\check{\alpha}$ of Beta distribution to 1.0 such that the interpolation coefficient $\lambda$ is uniformly distributed between zero and one.
We also analyze how different $\check{\alpha}$ will influence MetaMix.
The results are shown in Figure \ref{fig:beta_dist}.
We do the 5-way, 1-shot and 5-way, 5-shot tasks on {\it mini}-ImageNet and CUB datasets.
We can see that when $\check{\alpha}$ is below 1.0, the accuracy of both tasks on both datasets is a little lower.
When $\check{\alpha}$ is 1.0 and above, the performance maintains a good level.

\begin{figure}[b]
    \centering
    \includegraphics[width=8cm]{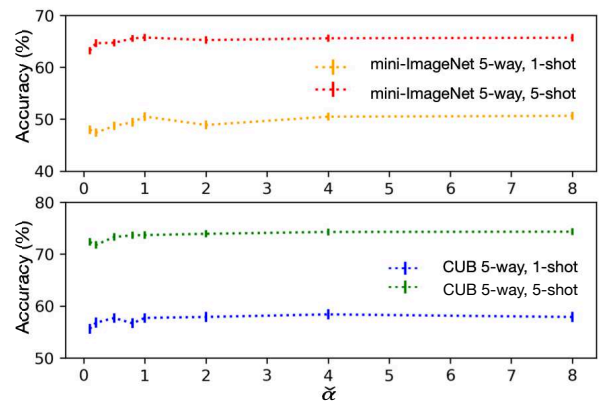}
    \caption{Effect of Beta distribution. $\check{\alpha}$ is set to 0.1, 0.2, 0.5, 0.8, 1.0, 2.0, 4.0, 8.0.}
    \label{fig:beta_dist}
\end{figure}

\newpage
\medskip
\noindent
\textbf{Effect of mixing on different sets}
\smallskip

\noindent
We make an ablation study here to compare the effect of doing mixup on different sets within an episode.
The results are listed in Table \ref{tab:ablation_study}.
We can see that mixing examples from the query set only performs best, while mixing examples from the support set only leads to a worse performance than the original MAML algorithm.
Moreover, mixing examples from both the support and query sets achieves a performance in between.

\begin{table}[h]
    \centering
    \begin{tabular}{lllll}
    \Xhline{2\arrayrulewidth}
    & \multicolumn{2}{c}{\textit{mini}-ImageNet} & \multicolumn{2}{c}{CUB} \\
     \multicolumn{1}{c}{Set(s)}    
    &\multicolumn{1}{c}{1-shot} &\multicolumn{1}{c}{5-shot}  &\multicolumn{1}{c}{1-shot} &\multicolumn{1}{c}{5-shot}  \\
    \hline
    \hline
     \multicolumn{1}{c}{$\mathcal{Q}$}
    &\multicolumn{1}{c}{\bf{50.51 $\pm$ 0.86}}
    &\multicolumn{1}{c}{\bf{65.73 $\pm$ 0.72}}
    &\multicolumn{1}{c}{\bf{57.70 $\pm$ 0.92}}
    &\multicolumn{1}{c}{\bf{73.66 $\pm$ 0.74}} \\
    \hline
     \multicolumn{1}{c}{$\mathcal{S}$}
    &\multicolumn{1}{c}{47.87 $\pm$ 0.82}
    &\multicolumn{1}{c}{62.34 $\pm$ 0.65}
    &\multicolumn{1}{c}{54.39 $\pm$ 0.97}
    &\multicolumn{1}{c}{67.23 $\pm$ 0.74} \\
     \multicolumn{1}{c}{$\mathcal{Q}$+$\mathcal{S}$}
    &\multicolumn{1}{c}{48.36 $\pm$ 0.81}
    &\multicolumn{1}{c}{64.06 $\pm$ 0.72}
    &\multicolumn{1}{c}{54.32 $\pm$ 0.93}
    &\multicolumn{1}{c}{70.30 $\pm$ 0.75} \\
    \Xhline{2\arrayrulewidth}
    \end{tabular}
    \caption{An ablation study of doing {\it mixup} on different sets. $\mathcal{Q}$ denotes the query set and $\mathcal{S}$ denote the support set.}
    \label{tab:ablation_study}
\end{table}

\medskip
\noindent
\textbf{Effect of the size of training data}
\smallskip

\noindent
Few-shot classification problems face the challenge that there are very few labeled examples of the novel classes, so that methods like meta-learning and transfer learning train on previous classes, which have large-scale data.

However, we are curious about what will happen if the previous classes do not have enough data.
MetaMix can be viewed as a technique of data augmentation, which plays an important role when training data are limited.
To verify the effectiveness of MetaMix from this aspect, we do another experiment to explore the impact of the size of training data on the performance of MetaMix.

We reduce 50\% of the training data and conduct experiments of MAML and MetaMix with MAML.
The results are shown in Table \ref{tab2:labelrate}.
We can see that MetaMix with MAML trained on 50\% data gets similar or even better performance, compared with MAML trained on 100\% data.
In the 5-way, 1-shot task, both MAML and MetaMix with MAML have a 1\% $\sim$ 5\% performance decrease on \textit{mini}-ImageNet and CUB.
However, in the 5-way, 5-shot task, MAML has a decrease of 3\% $\sim$ 6\%, while MetaMix with MAML has only 2.2 \% and 3.5 \% decreases on \textit{mini}-ImageNet and CUB, respectively.
As for the FC100 dataset, there are very small changes in both 1-shot and 5-shot tasks, which leads us to consider whether the meta-learning algorithms have learned enough information from the dataset.

Moreover, to make deeper analysis of the accuracy decreases caused by reducing the size of training data, we reduce 60\% and 70\% of the training data and do 1-shot and 5-shot classification on the \textit{mini}-ImageNet and CUB datasets.
The results are shown in Figure \ref{fig2:limitedlabel}.

\begin{figure*}[h]
\includegraphics[width=14cm]{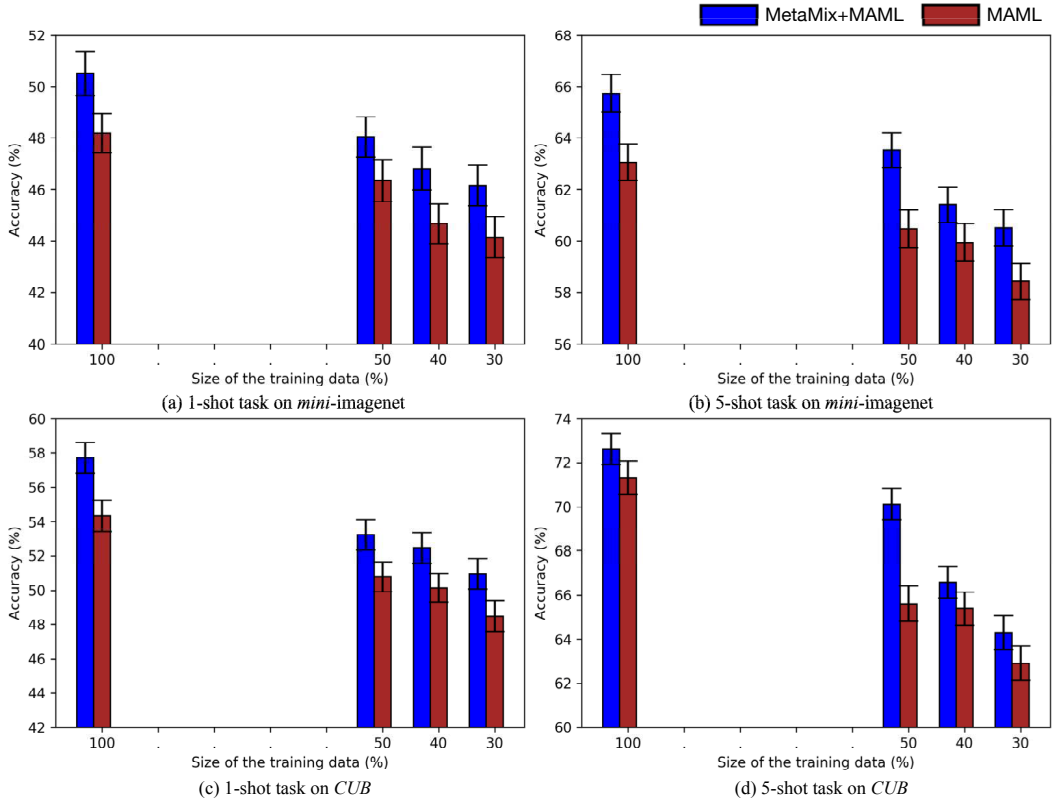}
\centering
\caption{A comparison among using 100\%, 50\%, 40\%, and 30\% of the training data: (a) 5-way, 5-shot task on \textit{mini}-ImageNet; (b) 5-way, 1-shot task on \textit{mini}-ImageNet; (c) 5-way, 1-shot task on CUB; (d) 5-way, 5-shot task on CUB.}
\label{fig2:limitedlabel}
\end{figure*}

We observe the accuracy change with the reduction of the training data size.
The accuracy and the size of training data are not linearly related.
In most cases, the accuracy drops faster with the reduction of the training data size, for both MAML and MetaMix with MAML.
We can see it clearly from the 1-shot task on the two datasets in diagrams (a) and (c).
But MetaMix with MAML has less performance decrease, compared with MAML, especially when cutting the amount of training data in half.
Moreover, we can see from diagrams (b) and (d) that when using MetaMix in the 5-shot task, there is a big accuracy drop between 50\% and 40\% training data size.
In other words, adding training data from 40\% to 50\% is much more effective than adding training data from 50\% to 100\%.
It makes us rethink the efficiency of using the training data.
Much more training data do not mean much better performance.

\newpage

\medskip
\noindent
\textbf{Discussion and future work}
\smallskip

\noindent
The MetaMix approach generates virtual examples upon the query set within each episode through a simple but effective way to regularize the backbone models.
By integrating it with MAML and its variants, we can improve the performance.
However, it also bring some questions we have mentioned above.

Firstly, we consider the poor performance in the FC100 dataset.
How to deal with data with low quality and how to transfer the initializer learned from one super-class to the other very different super-class?
For the former question, one possible answer is to use high-quality data from other collections and do cross-domain meta-learning.
For the latter question, a reasonable solution is to define a metric to evaluate the difficulty of a classification task, so that we can develop strategies which can learn dynamically to solve the tasks from easy to hard.

Then we think about the effect of the training data size.
From our empirical analysis, the performance of MetaMix does not drop too much, even when using only half of the training data.
It shows to a certain degree that MetaMix is a good data augmentation approach, but also prompts us to think about how to make better use of the information from the extra data.
In the future, we plan to adopt more knowledge from the training data and apply it to meta-learning.

Another question to be mentioned is about the interpolation-based consistency regularization.
According to our design, we do the interpolations on labeled examples from the query set.
Can we do interpolations on unlabeled examples or both the labeled and unlabeled examples to solve un/semi-supervied few-shot classification problems?
In our subsequent study, we will try more methods, such as \textit{mixMatch}.

\section{Conclusion}
\label{sec5:conc}

In this paper, we propose MetaMix as an approach to integrate with many meta-learning algorithms like MAML and its variants.
MetaMix produces virtual feature-target pairs in the query set within each training episode by a simple but effective interpolation-based method called \textit{mixup}.
Through this way, it learns an initializer which generalizes better to new tasks with a few fine-tuning steps.
Experiments on the \textit{mimi}-ImageNet, \textit{CUB}, and FC100 datasets demonstrate that MetaMix improves the performance of  MAML-based algorithms in few-shot classification tasks.
In the future, we will try to redesign the MetaMix framework to fit metric-based and model-based meta-learning algorithms.
We will further investigate utilizing more and better knowledge from the training data, which can be from both the same or different domains.
Moreover, we would like to apply our approach to broader areas, such as speech recognition and language processing.
\newpage
\bibliographystyle{IEEEtran}
\bibliography{MetaMix-ICPR}


\end{document}